\documentclass[letterpaper, 10 pt, conference]{ieeeconf}  

\IEEEoverridecommandlockouts                              

\usepackage[hyphens]{url}
\usepackage[english]{babel}
\usepackage[T1]{fontenc}
\usepackage[utf8]{inputenx}
\usepackage{graphics} 
\usepackage{epsfig} 
\usepackage{mathptmx} 
\usepackage{times} 
\usepackage{amsmath} 
\usepackage{amssymb}  
\usepackage{caption}
\usepackage{subcaption}
\usepackage{algorithm}
\usepackage[noend]{algpseudocode}
\usepackage{float}
\usepackage{wrapfig}
\usepackage{siunitx}
\usepackage{hyperref} 
\hypersetup{draft} 

\newcommand{\norm}[1]{\mathinner{\Vert#1\Vert}}%
\newcommand{\bmat}[1]{\begin{bmatrix}#1\end{bmatrix}}%

\graphicspath{{figures/}}

\title{\LARGE \bf
Non-Holonomic RRT \& MPC: Path and Trajectory Planning for an Autonomous Cycle Rickshaw
}

\author{Damir Bojadžić\authorrefmark{1}, Julian Kunze\authorrefmark{1}, Dinko Osmanković\authorrefmark{2}, Mohammadhossein Malmir\authorrefmark{3}, Alois Knoll\authorrefmark{3}
\thanks{\authorrefmark{1}WARP Student Team \& Department of Informatics, Technical University of Munich (TUM), Germany
    {\tt\small <damir.bojadzic, julian.kunze>@tum.de}}%
\thanks{\authorrefmark{2}Department of Automatic Control and Electronics, Faculty of Electrical Engineering, University of Sarajevo, Bosnia \& Herzegovina
    {\tt\small dinko.osmankovic@etf.unsa.ba}}%
\thanks{\authorrefmark{3}Chair of Robotics, Artificial Intelligence and Real-time Systems, Department of Informatics, Technical University of Munich (TUM), Germany
    {\tt\small hossein.malmir@tum.de, knoll@mytum.de}}%
}

\begin{document}

\maketitle
\thispagestyle{empty}
\pagestyle{empty}

\begin{abstract}

This paper presents a novel hierarchical motion planning approach based on Rapidly-Exploring Random Trees (RRT) for global planning and Model Predictive Control (MPC) for local planning. The approach targets a three-wheeled cycle rickshaw (trishaw) used for autonomous urban transportation in shared spaces. Due to the nature of the vehicle, the algorithms had to be adapted in order to adhere to non-holonomic kinematic constraints using the Kinematic Single-Track Model.

The vehicle is designed to offer transportation for people and goods in shared environments such as roads, sidewalks, bicycle lanes but also open spaces that are often occupied by other traffic participants. Therefore, the algorithm presented in this paper needs to anticipate and avoid dynamic obstacles, such as pedestrians or bicycles, but also be fast enough in order to work in real-time so that it can adapt to changes in the environment. Our approach uses an RRT variant for global planning that has been modified for single-track kinematics and improved by exploiting dead-end nodes. This allows us to compute global paths in unstructured environments very fast. In a second step, our MPC-based local planner makes use of the global path to compute the vehicle's trajectory while incorporating dynamic obstacles such as pedestrians and other road users.

Our approach has shown to work both in simulation as well as first real-life tests and can be easily extended for more sophisticated behaviors.
\end{abstract}


\section{INTRODUCTION}

In the context of autonomous driving, the focus usually lies on cars and trucks that can be used for the transportation of people and goods~\cite{SurveyAD,PlanningAV,AutonomousVehicleStack}. On the other hand, there are also small wheeled mobile robots that are employed for industrial intra-factory logistics and package delivery tasks~\cite{AutonomousMobileRobotsIntralogistics,MobileRobotsReview,LogisticsRobotsFramework}. While the former type usually operates under a set of traffic regulations and on a network of roads at velocities of sometimes more than $100\;km/h$, the latter is often required to just drive at a pedestrian speed and in contrast finds itself in more unstructured environments without a clear set of rules. However, our work targets a class of vehicles that is situated between a car and a small mobile robot, both with respect to the environment as well as the level of dynamics.\par
In this regard, the vehicle is designed to appropriately address passengers and goods transportation in shared urban environments like parks or bicycle lanes and also larger exhibition sites or factory premises. Rickshaws are far more space efficient than cars for single-person transportation. Combined with the reduced spatial requirements emerging from autonomous operation, this renders large parts of parking spaces unnecessary, which therefore allows cities to become more human-centered instead of vehicle-centered.\par

\begin{figure}[t]
    \centering
    \framebox[0.5\textwidth]{\parbox{0.485\textwidth}{\includegraphics[width=0.485\textwidth]{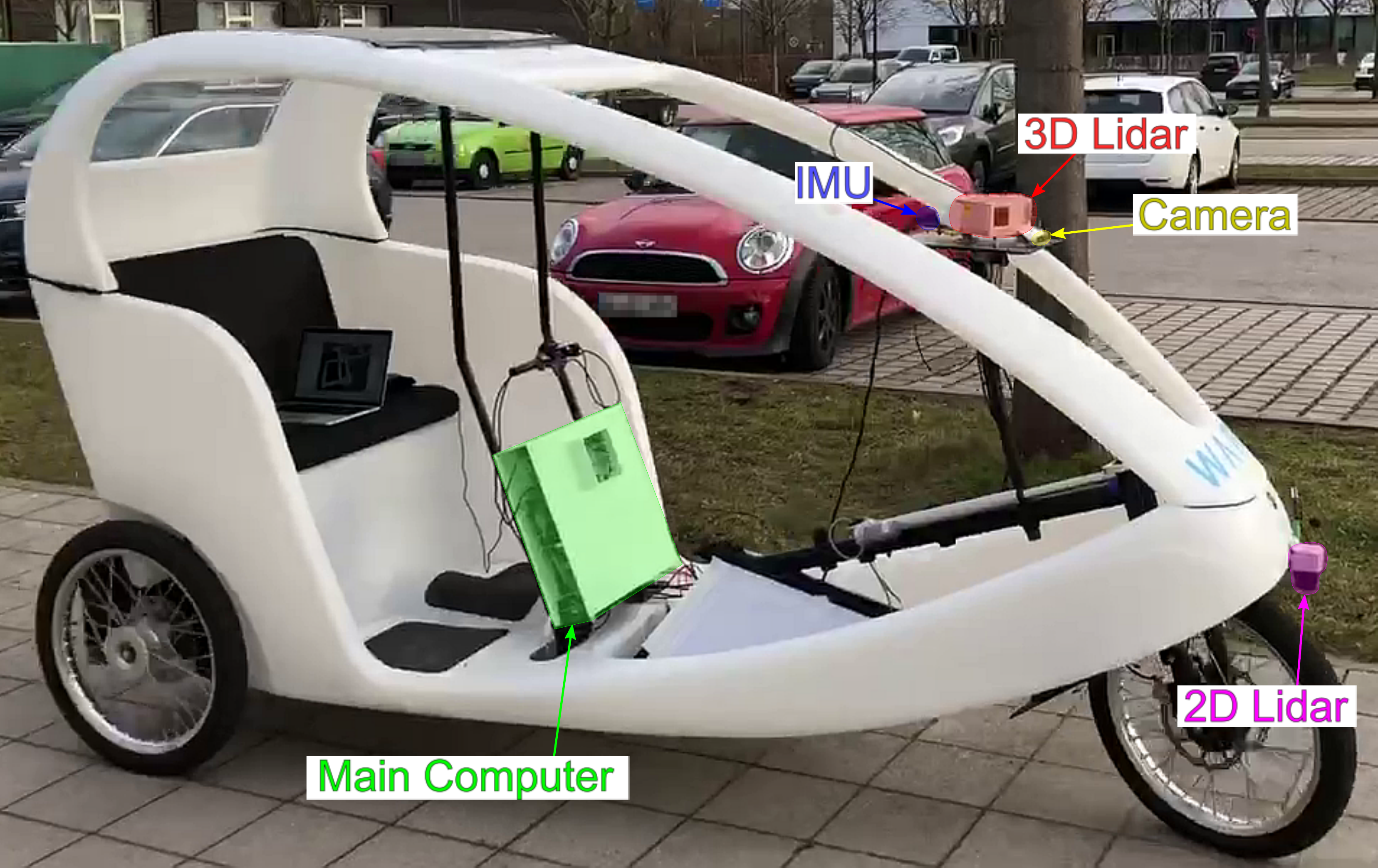}}}
    \caption{A photo of our target vehicle with main computer and retrofitted sensors highlighted. \vspace{-5mm}}
    \label{fig:rickshaw}
\end{figure}

In this paper, we want to tackle some of the significant challenges to the vehicle's motion planning that arise from autonomous driving in shared urban environments. Firstly, these environments require the vehicle to drive not only in a network of streets, but also in public open spaces. Secondly, sharing spaces with pedestrians, cyclists and other road users requires the planner to incorporate their behavior as well. Therefore, our approach consists of a two-step hierarchical planning architecture with an RRT-based path planner and an MPC-based trajectory planner. The planner is embedded into a system architecture based upon the Robot Operating System (ROS1)~\cite{ROS} and Kubernetes\footnote{\url{https://kubernetes.io/}}. Our RRT planner is able to traverse large open environments very quickly while taking care of the vehicle's kinematic constraints and to provide a global path on a static map for our vehicle to follow. The MPC planner computes the trajectory for following the global path but also incorporates dynamic obstacles like pedestrians. While in its current state, the MPC planner only avoids obstacles in a na\"ive manner, its architecture is designed to be able to add sophisticated behavior predictions for other road users easily.

\section{RELATED WORK}

\subsection{Rapidly-Exploring Random Trees}

Rapidly-Exploring Random Trees (RRT) count toward some of the most popular planning algorithms. Many modifications and improvements to the algorithm have been introduced. One of such is presented in \cite{7487439}, where the authors introduce the \textit{Theta$^{*}$-RRT} algorithm that hierarchically combines (discrete) any-angle search with (continuous) RRT motion planning for non-holonomic wheeled robots. The authors show that this variant outperforms its siblings \textit{RRT} \cite{lavalle1998rapidly}, \textit{A$^*$-RRT} \cite{6935304}, \textit{RRT$^*$} \cite{Gammell_2014} and \textit{A$^*$-RRT$^*$} \cite{6631372} both in path length and speed.

Another extension of RRT is presented in \cite{4018389}, where the authors suggest the usage of so-called key configurations for tackling problems such as moving through narrow passages with non-holonomic vehicles. 

\subsection{Model Predictive Control}

This paper was additionally inspired by the work of~\cite{Osmankovic2017AllTV}, where the authors utilized a Model Predictive Control (MPC) approach for local planning. Moreover, global path cost-to-go computation is done via a fast $D^*$-lite algorithm, and a Gaussian smoothing is applied on the precomputed costmap. The authors show that MPC can run on such a precomputed map and that MPC is robust enough to adapt to dynamic obstacles that appear at runtime that were not part of the original costmap. That way, the dynamic nature of the environment is also accounted for. 

In addition to the previous research paper,~\cite{nonlinearMPConline} represents a novel online motion planning approach based on nonlinear MPC. Using non-euclidean rotation groups, the authors have formulated an optimization problem that solves local planning via optimal control. The authors demonstrate that the proposed algorithm is able to solve a quasi-optimal parking maneuver whilst also avoiding dynamic obstacles. 

The power and capability of MPC extends even to racing scenarios. In~\cite{8493197} the authors propose a Linear Time Varying Model Predictive Control (LTV-MPC) approach for achieving minimum lap time in a racing scenario while also modeling the ability of the vehicle to drift in corners, thus giving it a time advantage. The MPC approach was not only able to keep up with the high demand and complex dynamic model of the vehicle, but also maintains stability under extreme drifting conditions. 

To the best of the authors knowledge, there is no existing prior work that suggests using Non-Holonomic RRT for global planning and online MPC for local planning designed for a three-wheeled cycle rickshaw. In the following sections, the vehicle and system architecture are first introduced followed by in-depth description of the adapted approach for path and trajectory planning steps.

\section{VEHICLE AND SYSTEM ARCHITECTURE}

Our approach targets a trishaw (three-wheeled cycle rickshaw) vehicle that has been modified to operate autonomously. The vehicle base is a BAYK CRUISER\footnote{https://bayk.ag/en/cruiser/} (see \autoref{fig:rickshaw}) where driver seat, handlebar, and pedal-powered drive have been removed. Instead, the rear wheel electric drive is controlled directly by our computer. Moreover, a separate steering motor and braking actuator have been installed to allow these functions to be computer-operated as well.\par
\begin{figure*}[t]
    \centering
    \framebox[\textwidth]{\parbox{0.99\textwidth}{\includegraphics[width=0.99\textwidth]{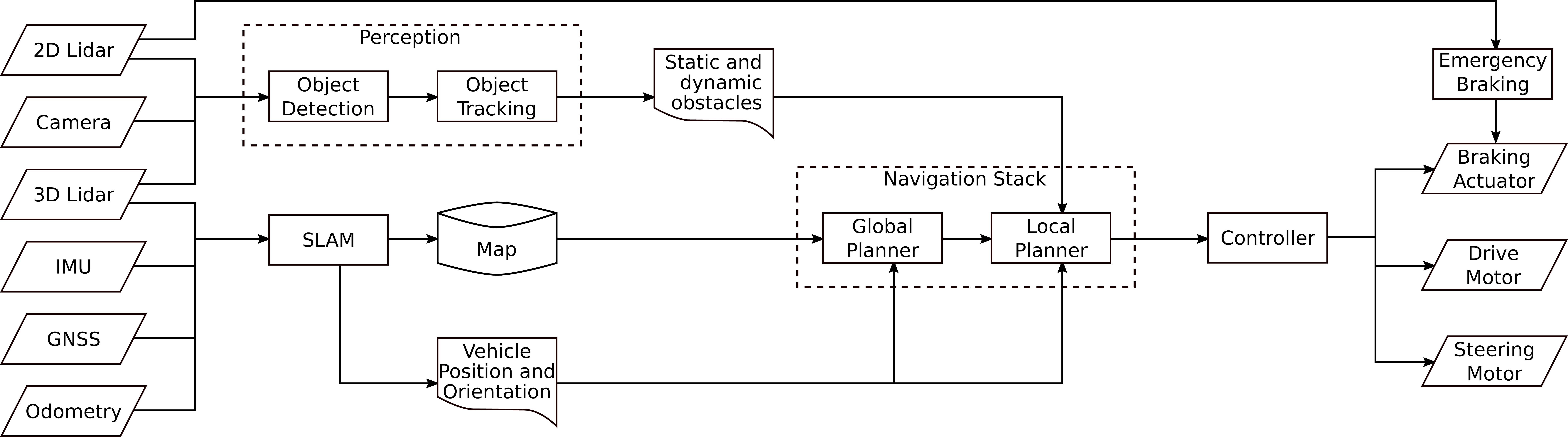}}}
    \caption{A simplified system architecture highlighting our vehicle's key components for locomotion. \vspace{-3mm}}
    \label{fig:sys_arch}
\end{figure*}

Besides the vehicle's proprioceptive sensors for measuring e.g. accelerations, wheel speed and steering angle, several exteroceptive sensors have been added in order to allow environmental perception and mapping. \autoref{fig:rickshaw} highlights the main computer and sensors that have been retrofitted to the vehicle. For triggering emergency braking, a $270^{\circ}$ FOV 2D Lidar has been mounted at the front. Furthermore, data from an RGB camera and a 3D Lidar is fused for object detection, providing information on static and dynamic obstacles. Finally, several proprioceptive sensors are combined with a GNSS sensor as well as the 3D Lidar for mapping the environment and localizing the vehicle on that map. The navigation stack uses processed obstacle information, map and vehicle pose to first generate a global path and then to iteratively compute a trajectory for the vehicle to follow. Finally, simple PID controllers for each separate actuator make sure that the trajectory is being correctly executed. \autoref{fig:sys_arch} gives an overview over the key components that are relevant for autonomously operating the vehicle.\par
Regarding our software architecture, all components have been containerized and are orchestrated via K3s Kubernetes\footnote{https://rancher.com/docs/k3s/latest/en/} in order to simplify a capsuled development of individual components as well as to allow for hot redundancy which is an important feature for safety critical systems~\cite{NilsKubernetes}. In addition, we employ ROS1\footnote{https://www.ros.org/} to facilitate the communication between individual components. Consequently, our navigation stack is based on the ROS navigation package\footnote{http://wiki.ros.org/navigation} of which we use the default map server but implement our own global and local planner. For mapping and localization, we modified the Google Cartographer ROS integration\footnote{http://wiki.ros.org/cartographer} to match our sensor setup.\par 
The next section will go into detail on how the global and local planner have been designed in order to match our use-case. 
\vspace{-3mm}

\section{APPROACH}
\label{sec:approach}


\subsection{GLOBAL PLANNER}

The global planner computes a global path $\mathcal{X}$ from starting state $[x_0, y_0, \theta_0]$ to goal state $[x_n, y_n, \theta_n]$ with non-predefined length $n$ on a given map. The global path is then defined as an array of individual states as shown in \autoref{eqn:globalpath}.
\vspace{-2mm}

\begin{equation}
    \mathcal{X} = \bmat{
    
    \begin{bmatrix}
    x_0 \\ y_0 \\ \theta_0
    \end{bmatrix}, & 
    
    \begin{bmatrix}
    x_1 \\ y_1 \\ \theta_1
    \end{bmatrix}, &
    
    \begin{bmatrix}
    x_2 \\ y_2 \\ \theta_2
    \end{bmatrix}, &
    
    ... , &
    
    \begin{bmatrix}
    x_n \\ y_n \\ \theta_n
    \end{bmatrix}
    }
\label{eqn:globalpath}
\end{equation}

RRT~\cite{lavalle1998rapidly} has been chosen as a suitable candidate to tackle the global planning task. The expanding tree is able to quickly navigate around the non-convex map and it presents a trade-off between exploration and exploitation with regard to its branches. The main idea of the algorithm is to begin at a starting configuration and expand that starting state in a tree like fashion by expanding edges of the tree inside the unoccupied space. At its core, the algorithm consists of two main phases: the sampling phase and the extension phase.

\begin{figure}[t]
    \centering
    \framebox[0.49\textwidth]{\parbox{0.49\textwidth}{\includegraphics[width=0.49\textwidth]{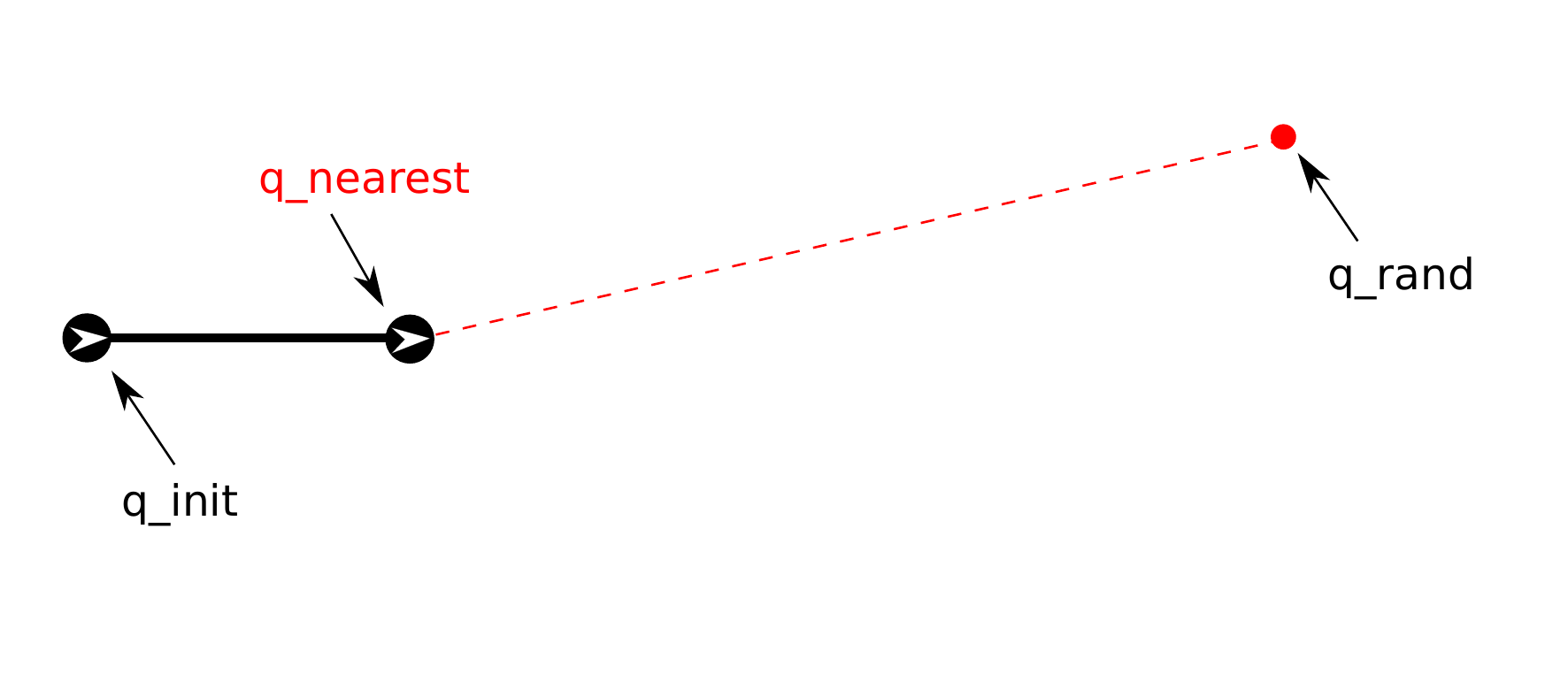} \\
    \includegraphics[width=0.49\textwidth]{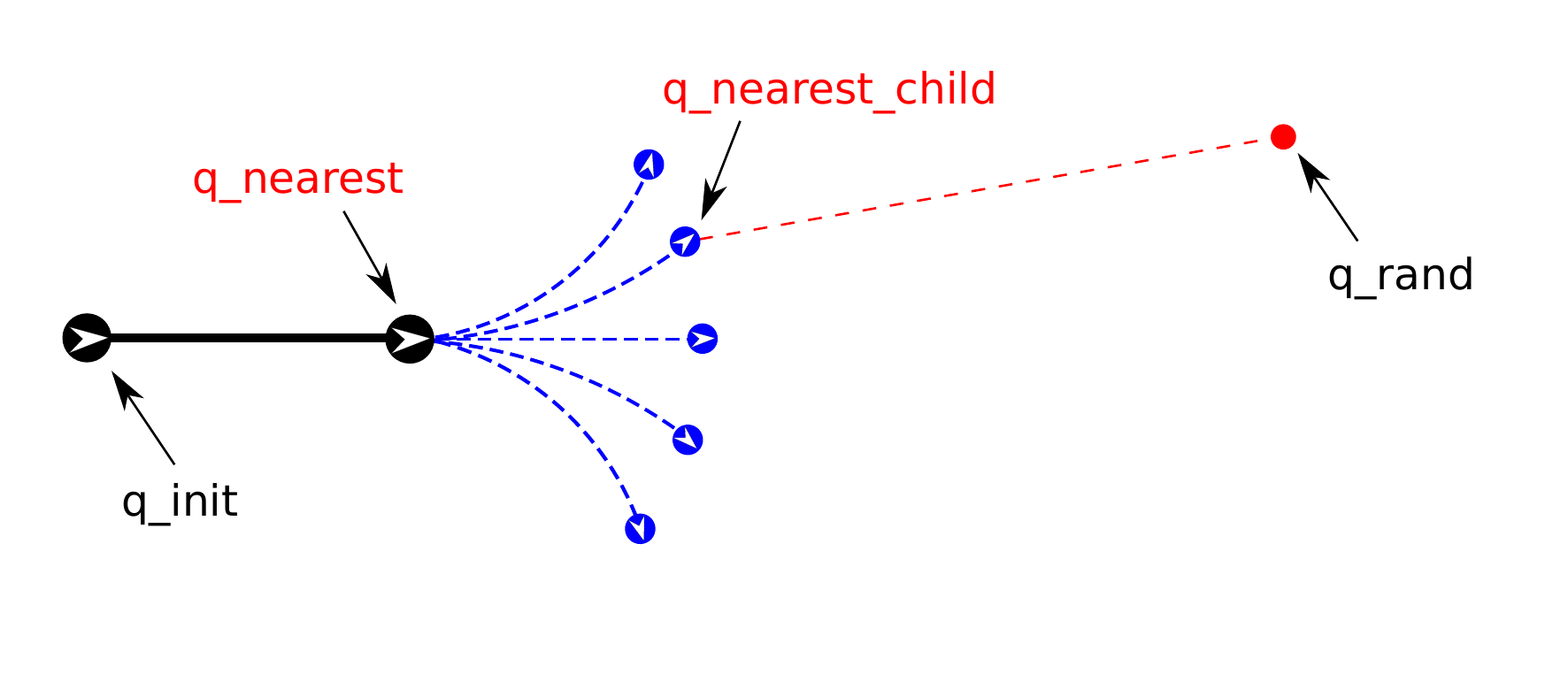}}}
    \caption{Non-holonomic RRT Algorithm: (top) Generating random point $q\_rand$ and finding the nearest node $q\_nearest$ inside the tree. (bottom) The nearest node is expanded into child nodes out of which the nearest is found and added to the tree.}
    \label{fig:nhrrt}
\end{figure}

In the sampling phase, a point $q\_rand$ is randomly sampled inside the unoccupied area of the map. After that, the state $q\_nearest$ is selected. This is illustrated in the upper part of \autoref{fig:nhrrt}. In the extension phase, the selected state is expanded in the direction of the randomly sampled point. This process is then repeated.

However, the vehicle in question adheres to non-holonomic single-track kinematic constraints, meaning that not all states that can be extended from the nearest state $q\_nearest$ are valid points. Therefore, the \textit{extend} strategy will have to be modified to incorporate non-holonomic constraints~\cite{rrtprogpros}, as defined by Equations \ref{eqn:kinematics_discrete_full} -- \ref{eqn:kinematics_discrete3}, where $\textbf{x}_k$ is the state of the vehicle at step $k$ and $\textbf{u}_k$ is the control input. The state vector $\textbf{x}$ is defined as $(x,y,\theta)$, with $x$ and $y$ given in $m$ and $\theta \in (-\pi, \pi]$ given in $rad$. The control input vector $\textbf{u}$ is defined as $(v, \delta)$, where $v$ is the linear velocity of the vehicle given in $m/s$ and $\delta$ is the front wheel steering angle in $rad$. $L$ is the length of the vehicle and $T$ is the discretized time step.
\vspace{-2mm}

\begin{alignat}{1}
    \label{eqn:kinematics_discrete_full}
    & \textbf{x}_k = \textbf{f}(\textbf{x}_{k-1}, \textbf{u}_k)\\
    \label{eqn:kinematics_discrete1}
    & x_k = x_{k-1} + v_k \cdot T \cdot \cos{\theta_{k-1}} \\
    \label{eqn:kinematics_discrete2}
    & y_k = y_{k-1} + v_k \cdot T \cdot \sin{\theta_{k-1}} \\
    \label{eqn:kinematics_discrete3}
    & \theta_k = \theta_{k-1} + v_k \cdot T \cdot \frac{\tan(\delta_k)}{L}
\end{alignat}

We utilize an approach similar to~\cite{Blanco2015TPSpaceR}  in order to incorporate the non-holonomic constraints into the RRT algorithm. To achieve this, the \textit{extend} phase will be modified such that a constant number of child-nodes is defined, that are generated using the non-holonomic equations, each with a steering angle ${\delta \in [min\_steering\_angle,\, max\_steering\_angle]}$ distributed evenly across the segment. The full extension strategy is defined in \autoref{alg:expansion} and illustrated in \autoref{fig:nhrrt}. Essentially, the non-holonomic constraints can be imagined as a cone of reachable states in front of the vehicle. 

\begin{algorithm}[t]
\caption{Extend Children}
\label{alg:expansion}
\begin{algorithmic}[1]
\State $children = \{\}$
\State $range = max\_steering\_angle - min\_steering\_angle$
\For{$i \leq steering\_samples$}
\State $\delta \gets  min\_steering\_angle + (i / steering\_samples) \cdot range$
\State $child \gets parent$
\For{$j \leq step\_size$}
\State $child.x \gets child.x + v \cdot \cos(child.\theta) \cdot T$
\State $child.y \gets child.y + v \cdot \sin(child.\theta) \cdot T$
\State $child.\theta \gets child.\theta + v / L \cdot \tan(\delta) \cdot T$
\State $j \gets j +  integration\_step\_size$
\EndFor
\State $children.insert(child)$
\State $i \gets i + 1$
\EndFor

\end{algorithmic}
\end{algorithm}

\begin{figure}[t]
    \centering
    \framebox[0.49\textwidth]{\parbox{0.49\textwidth}{\includegraphics[width=0.24\textwidth]{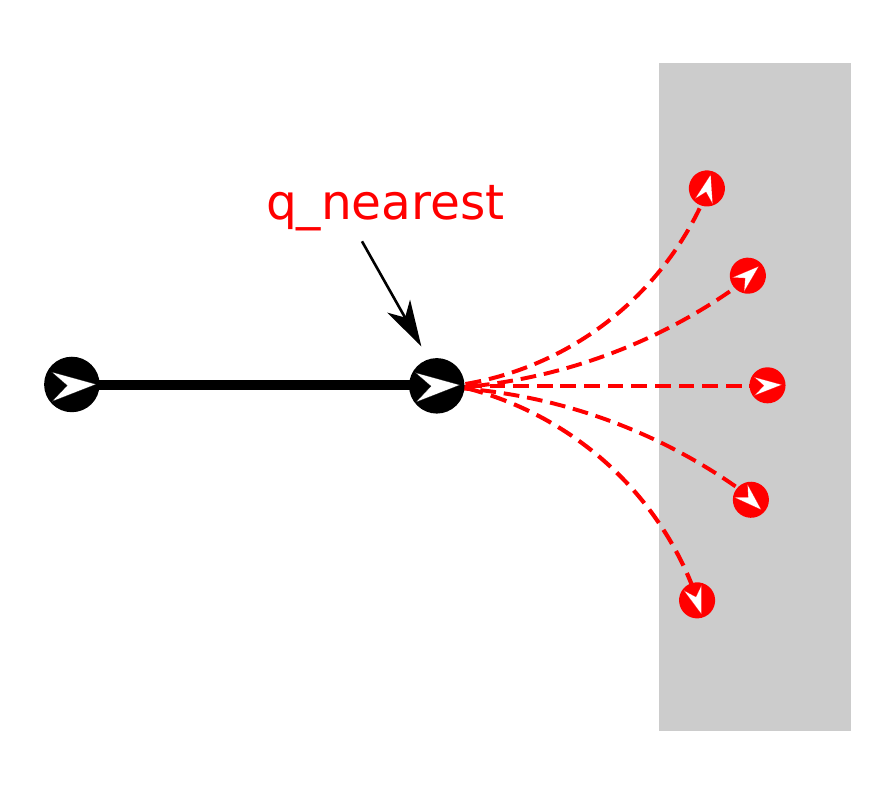} \includegraphics[width=0.24\textwidth]{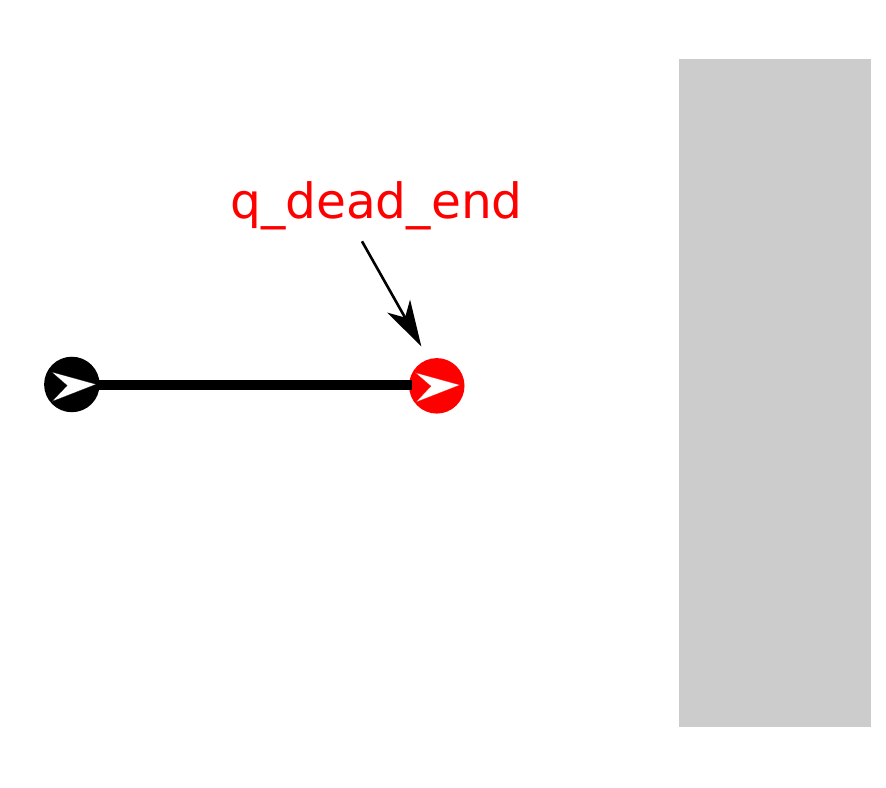}}}
    \caption{If a parent-node is unable to spawn child-nodes (left), that parent is marked as a dead-end node (right).}
    \label{fig:nhrrt2}
\end{figure}

Further modifications and improvements to the algorithm have been implemented. This paper introduces the concept of \textit{extended dead-end} nodes, which are nodes that cannot be further expanded. An example of these extended dead-end nodes can be found in \autoref{fig:nhrrt2}. The idea is as follows: If a node is unable to spawn any further child nodes \textit{and} all existing child nodes are marked as dead-end nodes, then that parent-node is also marked as a dead-end node. Nodes marked as dead-end are not considered for extension when searching the tree for $q\_nearest$ and therefore yield a speedup to the searching process. To the best of the authors' knowledge, this is the first time that such extension is applied to speed up the searching process in RRT path planners.

\textbf{Observation:} Due to the fact that the map is finite in size and a node can only be extended into a predefined constant number of child-nodes, the maximum number of possible nodes inside the tree is finite. Moreover, since in every iteration of the algorithm, a node is either added or removed (marked as dead-end) from the tree, an interesting consequence emerges: Given a starting pose $\textbf{x}_0$, the algorithm is guaranteed to cycle through every possible tree-state reachable from the starting position. This observation holds true regardless of the sampling strategy, even when we are sampling from the same point in every iteration.

\subsection{LOCAL PLANNER}

The task of the local planner is to compute a sequence of control signals $\textbf{u}$ that the vehicle has to follow. Besides knowledge of the map, the local planner also needs information about the dynamic obstacles inside the environment which the vehicle needs to avoid in order to not crash. The trajectory, as generated by the local planner, can be seen in \autoref{eqn:localtrajectory}:
\vspace{-2mm}

\begin{equation}
    \textbf{u} = \bmat{
    
    \bmat{v_0 \\ \delta_0 \\ \Delta\,t_0}, &
    \bmat{v_1 \\ \delta_1 \\ \Delta\,t_1}, &
    \bmat{v_2 \\ \delta_2 \\ \Delta\,t_2}, &
    ... , &
    \bmat{v_{n-1} \\ \delta_{n-1} \\ \Delta\,t_{n-1}}
    
    }
    \label{eqn:localtrajectory}
\end{equation}

The trajectory is a sequence of commands containing a linear velocity $v$ and a steering angle $\delta$. Furthermore, temporal information is embedded inside the trajectory as $\Delta\,t$, which represents the duration of the given control input. 

The sum of times $ T = \sum_i \Delta\,t_i $ is called the planning horizon and represents how far the algorithm will plan into the future. However, here the planner only returns and executes the first command of the sequence, i.e. $\mathbf{u} = [v_0, \delta_0]^T$ in each planning iteration. This in turn means that each command is then applied for a total time of $ \Delta\,t_0 = \Delta\,t_1 = \Delta\,t_2 = ... = \Delta\,t_{n-1} = \Delta\,t = 1 / f $ where $f$ is the constant frequency with which the planner is called.

Due to its ability to tackle optimization problems on a finite time-horizon, MPC has been chosen as a method for trajectory generation. The approach was inspired by~\cite{nonlinearMPConline}, builds on their ideas and takes advantage of their mathematical formulations.

The vehicle model is described by non-linear time-invariant differential equations with time step $k \in \mathbb{N}_0$, the state trajectory $\mathbf{x} : \mathbb{R} \mapsto \mathcal{X}$ and control trajectory $\mathbf{u} : \mathbb{R} \mapsto \mathcal{U}$, as given by Equations \ref{eqn:kinematics_discrete_full} -- \ref{eqn:kinematics_discrete3}. The mapping $\mathbf{f}: \mathcal{X} \times \mathcal{U} \mapsto \mathbb{R}^p$ with $p = dim(\mathcal{X})$ defines the change of state depending on control input, where $\mathcal{X}$ and $\mathcal{U}$ represent the state and control space respectively. The system is subjected to state and input constraints originating not only from the vehicle's physical limitations but also from the dynamic environment in which it drives. State constraints and input constraints will be introduced in the following.

The planning task is to guide the vehicle from $\textbf{x}_k \in \mathcal{X}$ at time step $k = 0$ to an intermediate or ultimate goal set $\mathbb{X}_f(t_N) \in \mathcal{X}$ within the planning horizon $T = N \Delta\,t$ while minimizing an objective function and adhering to constraints. 

With the running cost $\gamma: \mathcal{X} \times \mathcal{U} \mapsto \mathbb{R}_0^+$ and terminal cost $ \Gamma_f : \mathcal{X} \mapsto \mathbb{R}_0^+ $, the optimal control problem is given in \autoref{eqn:smallmpc}. For the intents and purposes of the local planner, the running cost will take into account factors such as trajectory length and distance from obstacles, whereas the terminal cost evaluates the final state $\mathbf{x}_N$ and its distance from the target point inside the global path. 
\vspace{-2mm}

\begin{equation}
    \min_{\textbf{u}}  \sum_{k=0}^{N-1} \gamma(\textbf{x}_k, \textbf{u}_k) \Delta\,t + \Gamma_f(\textbf{x}_N) \quad
    \label{eqn:smallmpc}
\end{equation}
\begin{alignat*}{2}
& \textrm{subject to:} \\
&    \textbf{x}_0 = \textbf{x}_s &
&    \textbf{x}_{k+1} = \textbf{f}(\textbf{x}_k, \textbf{u}_k) \\
&    \textbf{x}_k \in \mathbb{X}(k \Delta\,t) \quad &
&    \textbf{u}_k \in \mathbb{U}(\textbf{x}_k, k \Delta\,t) \\
&    T = N \Delta\,t & 
&    k = 0, 1, 2, ..., N-1 
\end{alignat*}

The constraint on the state $ \mathbf{x}_k \in \mathbb{X}(k \Delta\,t) $ implies that the vehicle should not find itself in an illegal state, for example colliding with other vehicles or static obstacles. The constraint is enforced as a penalty term inside the running cost function $\gamma(\cdot)$, but its validity is also manually checked once a trajectory has been chosen by limiting the obstacle costs to a maximum tolerated value. If that value is exceeded, the trajectory is rendered invalid. 

The constraint on the control inputs $\mathbf{u}_k \in \mathbb{U}(\mathbf{x}_k, k \Delta\,t)$ is derived from the physical real-world limitations of the vehicle and can be mathematically formulated as:
\vspace{-2mm}

\begin{equation}
    \begin{aligned}
        0 &\leq &&v_k & &\leq max\_velocity \\
        min\_steering\_angle &\leq &&\delta_k & &\leq max\_steering\_angle
    \end{aligned}    
\end{equation}

Since the aim of the local planner is to follow points on the global plan, a reasonable way to define the terminal cost function is to define a target point $\mathbf{x}_{target}$ on the global path and penalize the final state of the trajectory based on its distance from the target point. The terminal cost is therefore defined as shown in \autoref{eqn:terminalcost} where $c$ is a constant scaling factor for the penalization.

\begin{equation}
    \Gamma_f(\textbf{x}_N) = c \cdot \norm{\textbf{x}_{target} - \textbf{x}_N}_2^2
    \label{eqn:terminalcost}
\end{equation}

This definition of the terminal cost function makes intuitive sense, especially considering that the target points $\mathbf{x}_{target}$ moves along the global path as soon as the vehicle is within a certain range of it. This range is defined as the $global\_lookahead\_distance$. On the other hand, defining the running cost $\gamma (\cdot)$ on the other hand, involves three different terms and can be seen in \autoref{eqn:runningcost}:
\vspace{-2mm}

\begin{equation}
    \gamma(\textbf{x}_k, \textbf{u}_k) = c_l \cdot v_k \cdot \Delta\,t + c_m \cdot m(x_k, y_k) + c_o \cdot o(\textbf{x}_k)
    \label{eqn:runningcost}
\end{equation}

The terms $c_l$, $c_m$ and $c_o$ represent cost scaling factors and are of high importance for the planning process. Namely, the cost factors directly influence how much each of the penalty terms in the cost function influences the overall cost and therefore changes the shape of the final trajectory.

The first term of the running cost penalizes the length of the trajectory. This is done by penalizing the velocity as a higher velocity yields a longer trajectory.

The next term is penalization with respect to the map. The function $m(\cdot)$ evaluates the state based on the state coordinates $(x_k, y_k)$ and returns a discrete cost value between $[0, 255]$ where $0$ means that the space is free, and $255$ implies that the state is occupied. This value gradually decreases based on the distance from the obstacle, usually given by an inflation function. This inflation directly introduces a gradient that pushes the trajectory further away from the obstacle, by increasing running costs. The inflation is generated by the ROS map server package\footnote{\url{http://wiki.ros.org/map_server}} provided in the navigation stack.

The final term penalizes the state based on its relation to dynamic obstacles. This is done in a similar fashion to the way the previous term was evaluated. However, this term also offers the possibility of further extending the cost function with dynamic obstacle motion prediction. \autoref{eqn:obstacleprediction} offers a simple way of evaluating that problem, where M is the total number of obstacles.
\vspace{-2mm}

\begin{equation}
    o(\textbf{x}_k) = \sum_{i = 1}^{M} \phi(\norm{\textbf{obs}_i(k \Delta\,t) - \textbf{x}_k}_2)
    \label{eqn:obstacleprediction}
\end{equation}

The obstacle position is estimated at the time $k \Delta\,t$ and the distance is evaluated and penalized. The term $\mathbf{obs}_i(k \Delta\,t)$ is defined as the geometrical center of the obstacle $i$.

The distance to the obstacle is then passed to a function $\phi(\cdot) : \mathbb{R}_0^+ \mapsto \mathbb{R}$, which serves a similar purpose to the inflation function of the static obstacles and allows for adjusting the shape and intensity of obstacle inflation. The simplest way to define the function would be $\phi(x) = \frac{1}{x}$. This way, a cost gradient is introduced around each obstacle that penalizes closeness. This serves as a na\"ive first implementation and can be later adapted to suit the planning task best. For example, it makes sense to also incorporate a sense of motion prediction inside the function in order for the planner to anticipate movements of the obstacles.

In addition to these, further soft constraints can be imposed, such as penalization on spikes and jumps in the first derivative of the control input $\Dot{\mathbf{u}}$. For instance, introducing terms such as $\norm{\mathbf{u}_{k+1} - \mathbf{u}_k}_1$ penalizes variations in control inputs and yields a smoother trajectory. 

The next section deals with the evaluation of the suggested planning method.
\section{EVALUATION}
\label{sec:evaluation}

The evaluation of the approach is done in two stages, the first stage being inside a simulation environment and the second stage in the form of live tests on the vehicle. The optimal control problem is formulated and solved using Google's Ceres Solver \cite{ceres-solver}.

We conduct the simulation inside the Stage simulator ROS adaptation StageROS\footnote{\url{http://wiki.ros.org/stage_ros}}. The purpose of the simulations is to tune the planning parameters and evaluate the planner based on its ability to follow target points on a global path while avoiding static and dynamic obstacles. The dynamic obstacles represent pedestrians inside the map, which are simulated using the PedSim\footnote{\url{https://github.com/srl-freiburg/pedsim_ros}} simulator which uses the social force model for pedestrian dynamics~\cite{socialforce}.

The simulations have shown to be successful, with the vehicle reaching its target position while also managing to avoid dynamic obstacles. The global planner was able to generate paths of $\SI{150}\,{m}$ length in $\SI{0.2}\,{s}$ of computation time, whereas the local planner operates at a frequency of $\SI{5}\,{Hz}$ with a planning horizon of up to $T = \SI{12}\,{s}$ at times. These results have permitted to move ahead and test the approach on the target vehicle.

Over the period of a few months, several tests have been conducted live on the vehicle, with the most important improvements and highlights being recorded and uploaded to YouTube. In the real life tests, the planner has demonstrated its ability to calculate and follow a global path and reach the designated goal area while driving at a maximum speed of $\SI{1}\,{m/s}$. Due to the inflation of the static obstacles inside the map, the vehicle was able to keep a safe distance from walls and stay inside the designated driving area\footnote{\url{https://youtu.be/grg-59TbCVY}}. This holds true even when driving at night time\footnote{\url{https://youtu.be/Sm1tj7WEJLQ}}.

After that, further tests with dynamic obstacles were conducted. The vehicle is able to correctly identify pedestrians in its path and compute a trajectory that successfully avoids them. Using the 2D Lidar mounted in front of the vehicle, it is also able to detect immediate danger when a pedestrian jumps in front of the vehicle and to take immediate action by slowing down and turning right, thus avoiding collision with pedestrians\footnote{\url{https://youtu.be/Rnf0l_RLoyo}}.

Simulations and real life tests have shown that the suggested approach is valid and can be used for path and trajectory planning on the target vehicle. The vehicle and algorithm are still under development and are, as of the time of writing, not reliable enough for commercial use and fully autonomous driving. 

Testing has also revealed areas of potential improvement for the planner. One of those problems is that, even with inflation, the vehicle has a tendency to move very close to the edges of the driveable area and cut corners, causing the vehicle to freeze in front of obstacles with the global planner not able to find a path out from that state.

Another experienced issue arises from a lack of consistency when keeping up with the changing environment due to the dynamic nature of the obstacles. They sometimes cause the vehicle to come to an immediate stop (to avoid collisions with pedestrians) and request a re-planning of the global path. This can become problematic in crowds where numerous re-planning attempts would be made.

However, it is the authors' belief that these issues can be easily resolved by incorporating better obstacle inflation mechanisms, dynamic obstacle motion prediction functions and further hyperparameter tuning of the planner.

\section{CONCLUSIONS}

This paper presented a novel approach for online motion planning by utilizing a non-holonomic variant of the RRT algorithm for the purpose of global planning and an MPC approach for local planning designed for a three-wheeled rickshaw vehicle. Modifications have been made to both algorithms in order to better adapt them for the planning task. Among those modifications is the introduction of dead-end nodes in the non-holonomic variant of the RRT algorithm, as described in section \autoref{sec:approach}.

The planner performs well both in simulation and in real-life tests on the vehicle, as shown in \autoref{sec:evaluation}. It is able to generate a path from start to goal even in unstructured environments, and to follow that path to reach the goal region. Furthermore, the planner is able to avoid static and dynamic obstacles on its way and to keep a safe distance from walls and edges of the road by penalizing the vehicle's closeness to obstacles.

\addtolength{\textheight}{-12cm}   


\section*{ACKNOWLEDGMENT}

We would like to thank everyone involved in developing the entire hard- and software stack around this autonomous micromobility solution at the WARP student team.


\bibliographystyle{bibtex/bst/IEEEtranDOI}
\bibliography{IEEEabrv,references}

\end{document}